\documentclass[11pt]{article}

\usepackage[preprint]{acl}

\usepackage{times}
\usepackage{latexsym}
\usepackage[T1]{fontenc}
\usepackage[utf8]{inputenc}
\usepackage{microtype}
\usepackage{inconsolata}
\usepackage{graphicx}
\usepackage{booktabs}
\usepackage{multirow}
\usepackage{float}  %
\usepackage{amsmath}
\usepackage{amssymb}
\usepackage{url}
\usepackage{xspace}
\usepackage{enumitem}
\usepackage{xcolor}
\usepackage{eso-pic}

\makeatletter
\newif\ifpreprintbanner
\ifacl@finalcopy\ifacl@pagenumbers\preprintbannertrue\fi\fi
\makeatother

\renewcommand{\topfraction}{0.9}
\renewcommand{\bottomfraction}{0.4}
\renewcommand{\textfraction}{0.1}
\renewcommand{\floatpagefraction}{0.7}
\renewcommand{\dbltopfraction}{0.85}     %
\renewcommand{\dblfloatpagefraction}{0.65}

\newcommand{\valCoreDocs}{462{,}494}

\newcommand{\valDCLMmedEng}{0.0005}
\newcommand{\valDCLMmedNonEng}{0.29}
\newcommand{\valNemQmedEng}{0.49}
\newcommand{\valNemQmedNonEng}{0.37}
\newcommand{\valDCLMhcAll}{-0.083}
\newcommand{\valDCLMhcEng}{+0.064}
\newcommand{\valJQLnonEng}{+0.137}
\newcommand{\valFWtwoNonEng}{+0.186}

\newcommand{\valDCLMnemQ}{-0.459}

\newcommand{\valIccHostHi}{0.62}
\newcommand{\valIccHostLo}{0.43}
\newcommand{\valIccHostN}{532{,}628}
\newcommand{\valIccHostClf}{10}
\newcommand{\valHostDocs}{3.01}
\newcommand{\valYieldHClo}{0.7}
\newcommand{\valYieldHChi}{1.8}
\newcommand{\valYieldOracleLo}{4.0}
\newcommand{\valYieldOracleHi}{8.0}

\newcommand{\valHCstabLo}{0.851}

\newcommand{\valPRstabHi}{0.923}
\newcommand{\valSampleRhoKthree}{0.53}
\newcommand{\valSampleRhoKfive}{0.57}
\newcommand{\valDeepenMultLo}{1.7}
\newcommand{\valDeepenMultHi}{7.8}
\newcommand{\valRhoLead}{-0.08}
\newcommand{\valMissedBudgetM}{58}
\newcommand{\valGemShare}{62}
\newcommand{\valGemHosts}{369{,}534}
\newcommand{\valGemPctHosts}{6.5}
\newcommand{\valLeadHosts}{5{,}654{,}186}
\newcommand{\valLowCovShare}{50}
\newcommand{\valGemCellShare}{26}
\newcommand{\valMedQualLo}{20}
\newcommand{\valMedQualHi}{4}
\newcommand{\valGemRhoLo}{-0.151}
\newcommand{\valGemRhoHi}{-0.057}
\newcommand{\valGemCrawls}{6}
\newcommand{\valNemAgreeLo}{0.53}
\newcommand{\valNemAgreeHi}{0.54}
\newcommand{\valFWeduAgree}{+0.29}
\newcommand{\valDCLMagree}{-0.06}
\newcommand{\valCovBlindMax}{0.07}
\newcommand{\valUKbudget}{23}
\newcommand{\valUKhosts}{6.1}
\newcommand{\valUKcov}{34}

\newcommand{\HC}{\textsc{HC}\xspace}
\newcommand{\PR}{\textsc{PR}\xspace}
\newcommand{\FWEdu}{FineWeb-Edu\xspace}
\newcommand{\DCLM}{DCLM\xspace}
\newcommand{\CFLLM}{Craw4LLM\xspace}
\newcommand{\qdens}{quality density\xspace}

\title{Host-Aware Web Crawling for LLM Pretraining Data}

\author{Malte Ostendorff \\
  Common Crawl Foundation \\
  \texttt{malte@commoncrawl.org}}

\begin{document}

\maketitle

\begin{abstract}

Web crawling is a major source of large language model (LLM) pretraining data, yet data-selection pipelines discard a large share of crawled pages. We investigate how a crawler should allocate its budget across and within hosts to collect more high-quality pretraining data. Combining document-level quality classifier scores, web-graph centrality metrics, and Common Crawl's per-host crawl coverage, we find that roughly half of the variation in classifier scores occurs between hosts. In the crawls we analyze, a few pages from a host also provide a useful estimate of the quality of its remaining pages. Common Crawl uses harmonic centrality, a graph centrality metric, to guide its crawl allocation. In our simulations, using centrality to choose which hosts to crawl yields limited gains over random ordering and performs worse for some classifiers. For deciding how deeply to crawl each host, sampling a few pages and then fetching more from hosts with higher-scoring samples collects several times more high-quality pages per fetch than uniform depth or centrality-based allocation. Finally, Common Crawl's observed per-host coverage has little correlation with the evaluated quality signals. We estimate that tens of millions of high-quality pages remain among known but uncrawled URLs in each crawl examined, with most concentrated on a small share of hosts.

\end{abstract}

\section{Introduction}
\label{sec:intro}

Large language model (LLM) pretraining relies heavily on Web data such as the Common Crawl (CC) corpus, which is collected by
crawling. A Web crawler with a finite budget, like Common Crawl's crawler, must decide continuously \emph{which} URLs
to fetch next \citep{cho1998efficient}.
CC decides this primarily based on a Web graph metric, namely harmonic centrality \citep[HC;][]{boldi2014axioms}.

Because most crawled pages are later discarded as low quality by pretraining pipelines
\citep{li2024datacomp,albalak2024survey}, the crawl policy decision can be seen as a
pretraining data selection or quality decision. 
However, the crawling aspect of pretraining data has been mostly neglected in the literature.
Only recently, \citet{yu2025craw4llm} proposed with \textit{Craw4LLM} a policy that prioritizes the crawl frontier by the
pretraining-classifier score of each fetched page, reaching parity with a full crawl from
a fraction of the URLs.

Such page-level schedulers are host-blind: they maintain a global per-page priority
queue and must fetch, extract, and classify \emph{every} candidate page to obtain the
scheduling signal. 
We ask a complementary question at the host level, one that requires far less compute: given the Web graph and data-selection classifiers, how should a crawler allocate its budget across and within hosts - and is Common Crawl's deployed allocation already aligned with what pre-training quality selection would choose?
To answer it, we study ten data-selection classifiers across several crawls and how they correlate with Web graph metrics and Common Crawl's own per-host crawl budget.
The paper is organized around four research questions that lead to the host-level crawl policy.

\paragraph{(RQ1)~Is document quality a host-level property?} A host-level
variance decomposition shows quality is moderately concentrated by host (ICC1
\valIccHostLo{}--\valIccHostHi{}), motivating our host-aware policy and providing the  %
per-host sample size it needs (\S\ref{sec:rq1}). 

\paragraph{(RQ2)~Do graph-centrality metrics work
as a crawl-yield signal?} A crawl-yield evaluation confirms and \emph{explains} the
negative graph-prior result of \citet{yu2025craw4llm} through a language confound and
heterogeneous hub composition, across ten classifiers (\S\ref{sec:rq2}). 

\paragraph{(RQ3)~How deep should a crawler go per host, and can a small sample decide?} A host-aware
crawl-\emph{depth} analysis shows a $\sim$3-page sample predicts a host's remaining
quality, and a sample-then-deepen policy approaches the oracle policy while centrality-biased depth does
not (\S\ref{sec:rq4}).

\paragraph{(RQ4)~Does Common Crawl's actual budget align with quality?} An analysis of historical crawl data finds that CC's per-host budget is uncorrelated with quality under every classifier we test, leaving tens of millions of known
high-quality URLs uncrawled on identifiable hosts (\S\ref{sec:rq5}).

\section{Related Work}
\label{sec:related}

\paragraph{LLM Pretraining Data.} 
Pretraining datasets are typically filtered using methods that
can process large volumes of text efficiently. These include
hand-written rules based on document length, symbol frequency,
stopword frequency, or language, as well as learned classifiers
that estimate each document's quality from its text.
The prevailing choice is the efficient fastText text classifier \citep{joulin2017bag},
trained on high-quality reference corpora in English: \DCLM \citep{li2024datacomp} uses a fastText
OH-2.5+ELI5 classifier, GneissWeb \citep{ibm2025gneissweb} an ensemble of five
topic-specialized fastText classifiers, and Nemotron-CC \citep{parmar2024nemotron} a
three-way fastText ensemble bucketed into five quality tiers; FineWeb-Edu
\citep{penedo2024fineweb} uses a more compute-intensive Transformer model. The same recipe extends across
languages: JQL-Edu \citep{ali2025jql} for over 35 languages,
and the FineWeb2-HQ classifier \citep{messmer2025multilingual} for the multilingual FineWeb2 corpus
\citep{penedo2025fineweb2}. Large open pretraining corpora such as Dolma
\citep{soldaini2024dolma} combine such signals into full, reproducible curation pipelines. All of these signals
are content-internal, since they look only at the text and not at Web graph or host-level features.
The only graph-adjacent signals in production pipelines are categorical domain
blocklists and the social-graph seeds \citep{gokaslan2019openwebtext, brack-etal-2024-community}.

\paragraph{Graph Signals in Web IR.}
Classical link-analysis methods include PageRank \citep{page1999pagerank},
HITS \citep{kleinberg1999authoritative}, harmonic centrality \citep{boldi2014axioms},
TrustRank \citep{gyongyi2004combating}, and anti-TrustRank. \citet{castillo2007know}
showed that link-topology features classify Web spam better than content alone.
CrediBench \citep{zhang2025credibench} trains a graph neural network on a
45M-node CC subgraph for source credibility prediction. LinkBERT \citep{yasunaga2022linkbert}
mines QA pairs from Wikipedia hyperlink graphs for synthetic pretraining data.
Both use graph structure for purposes other than document filtering.

\paragraph{URL-level Crawl Policy.} \label{sec:craw4llm} 
\textit{Craw4LLM} \citep{yu2025craw4llm} is the closest prior work. On ClueWeb22
\citep{overwijk2022clueweb2210billionweb}, the authors measure a weak Spearman $\rho \approx -0.11$
between a page's indegree and its \DCLM score (on the ClueWeb22-B sample), and simulate crawling on
the 900M-page ClueWeb22-A subset. Indegree-prioritised crawling reaches only 69.5\% of
oracle downstream LM performance, vs.\ 78.1\% for random crawling and 95.3\% for
fasttext-driven crawling. They conclude that graph centrality is a poor standalone
LLM data selector.
Our results reproduce the negative overall correlation ($\rho = -0.119$ for \HC on
CC-MAIN-2024-10) and provide a plausible mechanism: English-trained classifier
scores covary with host language composition, which itself covaries with \HC.
We do not analyse not-public ClueWeb22-A directly. 
Furthermore, we also find that \PR is positively correlated with quality when comparing hosts within the same top-level domain (TLD).

\paragraph{Graph-based Data Selection.}
\label{sec:webgraphmix}
\textit{WebGraphMix} \citep{badoni2026webgraphmix}  evaluates mixtures of already-collected documents for
LLM pretraining. We evaluate host ordering and depth allocation during
crawling, measuring high-quality pages collected under a fixed fetch
budget. We focus on harmonic centrality (\HC) and PageRank (\PR),
the metrics Common Crawl publishes. WebGraphMix's finding that
\PR-based selection does not improve downstream performance is
consistent with the limited crawl-yield gains from centrality-based
ordering in our experiments (\S\ref{sec:rq2}), although the two
studies evaluate different setups.
\DCLM \citep{li2024datacomp} grouped documents into five bins
by \PR and found that no bin outperformed random sampling.
Our results suggest a possible explanation: pooling hosts
across TLDs can obscure the positive correlation between
\PR and quality within individual TLDs.

\section{Data and Methods}
\label{sec:methods}

Our experiments draw on \emph{two distinct data regimes}, and keeping them separate is
important for interpreting the results. In the \textbf{classified-sample} regime we do our
own extraction and classification: we read WARC records, extract main text, 
and run our data-selection classifiers on documents extracted
  from 224 WARC files, approximately 0.25\% of the WARC files
  in CC-MAIN-2024-10 (RQ1--RQ4).\footnote{Our own classification is needed since pretraining datasets are typically published without low quality samples and without the corresponding full range of classifier scores.} 
In the \textbf{full-crawl} regime, we instead use precomputed artifacts (Common Crawl's
host index and the GneissWeb quality annotations covering \emph{every} fetched page) without
re-classification (RQ3 replication, RQ4). Appendix~\ref{app:datasets} lists every dataset
and model and which research question uses it.

\subsection{Classifying Web Documents}

Our primary classified sample is from CC-MAIN-2024-10 (February/March 2024).
Specifically, we extract the text from HTML pages using Trafilatura \citep{barbaresi-2021-trafilatura} from 224 randomly selected WARC files.
Keeping hosts with at least three documents yields
\valHostDocs{}M documents on \valIccHostN{} hosts for our  %
primary host-level analyses. These document counts refer to
our sample of WARC files, not to all pages that Common Crawl
has fetched from each host. The depth analysis uses stricter
inclusion thresholds, described in \S\ref{sec:rq4}.

We score each extracted document with ten fasttext classifiers and take host-median scores: \DCLM \citep{li2024datacomp}, \FWEdu \citep{penedo2024fineweb}, three
Nemotron-CC classifiers (quality, educational, information) \citep{parmar2024nemotron},
and five GneissWeb classifiers (a quality annotator plus science, education, technology,
and medical topic scorers) \citep{ibm2025gneissweb}.
Six of the ten classifiers target pretraining \emph{quality} and four score \emph{topic}. 
Throughout, we use
\emph{quality} to mean a classifier's predicted usefulness for pretraining, which is itself only a proxy for
downstream benchmark performance and not intrinsic document quality. 
For \FWEdu, we run a
compute-efficient fastText re-implementation \citep{ktsui2024fineweb} of the original
Transformer-based classifier, at comparable accuracy. 

For the language analysis
(\S\ref{sec:rq2}), we additionally run two multilingual Transformer classifiers, JQL-Edu
\citep{ali2025jql} and the FineWeb2-HQ classifier \citep{messmer2025multilingual,penedo2025fineweb2}, 
on a smaller core set of \valCoreDocs{} documents derived from 16 WARC files.  %
Exact model identifiers are in Table~\ref{tab:models}. 
These  scores drive the variance decomposition (RQ1), the crawl-yield simulation (RQ2), and the
primary depth simulation (RQ3). We run the same extraction-and-scoring pipeline on a
second snapshot, CC-MAIN-2023-40 (32 WARC files), which we use for the language and
cross-crawl robustness checks (Appendix~\ref{app:datasets}).

\subsection{Web Graph}

Graph features come from Common Crawl's public host-level hyperlinkgraph\footnote{\url{https://commoncrawl.org/web-graphs}}, using the
aligned \texttt{cc-main-2023-24} release (built from crawls spanning September 2023 to
February 2024) so that a single graph covers our document snapshots. For each host we
take harmonic centrality (\HC) and PageRank (\PR), the two centrality measures Common
Crawl publishes, in $\log_{10}(1{+}x)$ form. Fewer than 0.3\% of our hosts are missing
from the graph and are excluded.

\subsection{Crawl Budget and Quality Density}

The full-crawl regime uses two Common Crawl artifacts we do \emph{not} recompute
ourselves. The \textbf{host index} reports, per host, the number of known URLs
(\texttt{nutch\_numRecords}), how many were fetched (\texttt{nutch\_fetched}), and the
resulting \emph{crawl budget or coverage} (\texttt{nutch\_fetched\_pct}). The \textbf{GneissWeb
annotation set} \citep{ibm2025gneissweb}, released by Common Crawl, labels the pages a
strong quality filter kept; every annotated page passed the filter, and the annotations
cover \emph{every} fetched page rather than our WARC sample. Joining the two on the same
crawl gives a per-host \emph{\qdens},
\begin{equation}
\text{\qdens} = \frac{\#\text{\small GW-quality pages}}{\#\text{\small fetched pages}},
\end{equation}
the fraction of a host's crawled pages that are high quality. Because it is full-crawl and
independent of our own extraction and classifiers, it both replicates the depth analysis
at full scale (RQ3) and grounds the crawl-budget analysis (RQ4). 
We validate it against
our own classifiers in \S\ref{sec:rq5}. 
Concretely, the crawl-budget analysis joins the
host index with GneissWeb quality annotations on CC-MAIN-2024-10 and also across six
crawls for the temporal panel (CC-MAIN-\{2021-49, 2022-49, 2023-23, 2023-40, 2024-10, 2024-18\}). 
The full-crawl depth replication uses the GneissWeb depth
annotations for CC-MAIN-2017-34.

\subsection{Metrics}

We use different metrics
for different research questions. RQ1 examines variation in individual
page scores. The graph-quality correlations use the median
classifier score of each host. The crawl-yield and depth
simulations instead count pages whose classifier score exceeds
a threshold and measure the proportion of high-quality pages.
For RQ4, quality density is the proportion of fetched pages
retained by the GneissWeb filter. 
A host's median score and its
proportion of high-quality pages are different quantities.

\paragraph{Spearman $\rho$.} Our primary statistic is Spearman rank correlation, which
measures whether two quantities tend to rise together ($+1$) or fall as the other rises
($-1$), with $0$ meaning no monotone relationship.

\paragraph{Host-level variance (RQ1).} The intraclass correlation (Shrout--Fleiss one-way
estimator)
\begin{equation}
\text{ICC1} = \frac{\sigma^2_\text{between}}{\sigma^2_\text{between}+\sigma^2_\text{within}}
\end{equation}
is the share of per-document score variance that lies \emph{between} hosts. ICC1 $=1$ would
mean a host's pages are uniform in that score and $0$ that a host tells us nothing about a
page. With $\bar m$ documents per host, the Spearman--Brown reliability
\begin{equation}
\frac{\bar m\,\text{ICC1}}{1+(\bar m-1)\,\text{ICC1}}
\end{equation}
bounds the host-level correlation any signal can achieve, so ICC1 also sets the per-host
sample size needed to rank hosts.

\paragraph{Crawl-yield simulation (RQ2).} We fix a crawl budget as a fraction of corpus
documents and fetch hosts in the order a policy defines until the budget is spent. The policies are random, Common Crawl's
current allocation, \HC, \PR, or an oracle that knows each host's true quality
density. We report yield as the share of fetched pages that are
high quality (Figure~\ref{fig:rq2a}) and, for cross-classifier comparison, as a multiple of
the random baseline (Appendix~\ref{app:rq2_yield}).

\paragraph{Depth simulation (RQ3).}  We count a page as \emph{high quality} when its quality score exceeds $0.5$, and a host's \qdens{} is its fraction of such pages, using both our \FWEdu{} scores on the classified sample and GneissWeb quality scores on the full crawl. RQ1 already implies that a small per-host sample characterizes a host, but we also confirm this directly by correlating a $k$-page sample's \qdens{} with that of the host's remaining pages (Figure~\ref{fig:rq4a}), and then use the sample to decide crawl depth.

Each policy gets the same budget: a fixed number of pages to fetch, set as a fraction of all available pages. We score a policy by its \emph{yield}, i.e., the high-quality pages it collects per page fetched. A crawler cannot tell a host's high quality pages from its low quality ones until it fetches them. So the expected high-quality count is the host's quality density times the number of pages fetched.
A policy therefore cannot cherry-pick high quality pages \emph{within} a host. Instead, it can only choose \emph{which} hosts to crawl deeply. We compare four policies. 

\emph{Uniform breadth} crawls every host to the same shallow depth. \emph{\HC depth} crawls hosts fully in order of harmonic centrality, highest first, until the budget is spent. \emph{Sample-then-deepen} first probes three pages per host, then crawls fully the hosts whose sampled \qdens{} is the highest. The \emph{oracle} crawls fully the hosts with the highest \emph{true} \qdens{} first, giving the upper bound.

\paragraph{Crawl-budget analysis (RQ4).} We correlate each host's
crawl coverage with its quality density, and define \emph{hidden
gems} as hosts below the median coverage whose quality density
is in the top decile. We estimate the number of high-quality
pages among a host's known but uncrawled URLs by multiplying their
count by its observed quality density. This assumes that
uncrawled pages are as likely to be high quality as the fetched
pages from the same host.

\section{Results}
\label{sec:results}

\subsection{RQ1: Quality Is a Host-Level Property}
\label{sec:rq1}

\begin{figure}[tbp]
\centering
\includegraphics[width=\columnwidth]{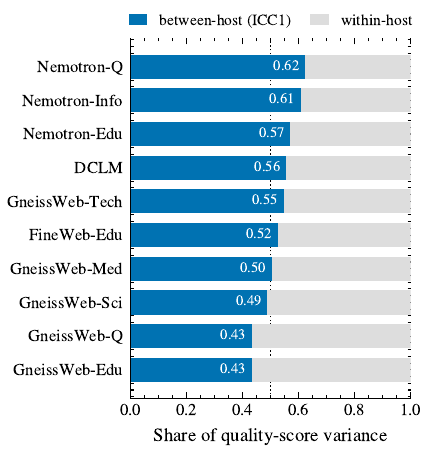}
\caption{The intraclass correlation ICC1 is the share of a classifier's per-document score
variance that lies \emph{between} hosts rather than within them: Across
all ten classifiers ICC1 is \valIccHostLo{}--\valIccHostHi{} ($N = \valIccHostN{}$ hosts), so  %
roughly half of the variation in page quality is between hosts. This suggests that a few sampled pages already characterize a host, enabling a host-level crawl policy.}
\label{fig:rq1}
\end{figure}

Allocating more crawl budget to some hosts can improve yield
if their pages tend to be higher quality than those on other hosts.
We therefore measure how much of the variation in classifier
scores occurs between hosts.
Across all \valIccHostClf{}  %
classifiers, the between-host share of per-document score
variance (ICC1) lies in \valIccHostLo{}--\valIccHostHi{}, indicating moderate host-level  %
  consistency (Figure~\ref{fig:rq1}). In other words, roughly half of the observed variation
in classifier scores is between hosts.

Moderate ICC1 has two consequences we use later. With $\bar m \approx 5$ observed documents per host in our WARC sample and ICC1 $\approx 0.5$, Spearman--Brown reliability is $\approx 0.83$, so the
largest host-level correlation any signal can reach is $\approx 0.9$, not $1$; near-zero
correlations must be read against this ceiling. 
The observed host-level consistency also motivates testing
whether a \emph{small} sample predicts the quality of remaining pages,
which we evaluate directly in RQ3.
  
\subsection{RQ2: Graph Centrality Is a Weak Crawl-Yield Signal}
\label{sec:rq2}

\begin{figure}[tbp]
\centering
\includegraphics[width=\columnwidth]{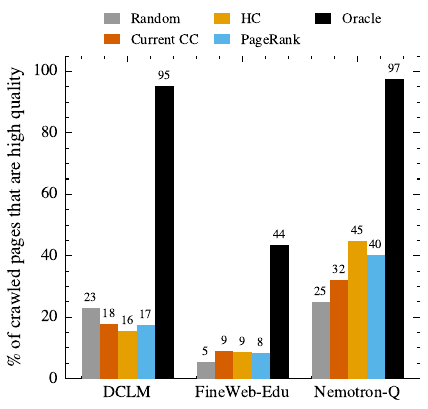}
\caption{Graph centrality is a weak signal for \emph{which} hosts to crawl. When only 10\%
of pages are crawled, the share of fetched pages that are high quality: centrality ordering
(current/\HC/\PR) captures only a fraction of the \emph{oracle} (crawler knows host's
quality), and even trails random for \DCLM.}
\label{fig:rq2a}
\end{figure}

Common Crawl uses harmonic centrality (\HC) to guide which hosts
to crawl and how deeply to crawl them. We therefore test whether
prioritizing hosts by centrality increases the number of high-quality
pages collected under a fixed crawl budget.

At a 10\% crawl budget, ordering hosts by \HC
collects \valYieldHClo{}--\valYieldHChi{} times as many high-quality  %
pages as random ordering across classifiers, compared with
\valYieldOracleLo{}--\valYieldOracleHi{} times for an oracle that  %
knows each host's quality density (Figure~\ref{fig:rq2a}).
PageRank (\PR) and Common Crawl's current allocation also yield
substantially fewer high-quality pages than the oracle.
For \DCLM, centrality-based ordering performs worse than random.
Ordering by \HC also roughly halves the number of top-level
domains represented compared with random ordering, reducing
TLD diversity. Appendix~\ref{app:rq2_yield} reports results
by classifier and crawl budget.

\begin{figure}[tbp]
\centering
\includegraphics[width=\columnwidth]{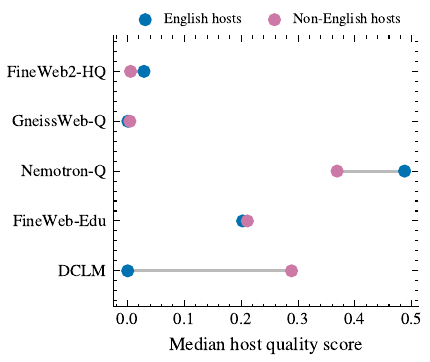}
\caption{
Median host quality scores by language. English-trained
classifiers differ in how they score English and non-English hosts:
\DCLM assigns higher scores to non-English hosts, while Nemotron-Q
assigns higher scores to English hosts. Because centrality also varies
by language, its correlation with classifier scores partly reflects
these language differences.}
\label{fig:rq2b}
\end{figure}

\paragraph{Language partly explains the weak correlation.} Prior work reports the similar findings.
\CFLLM measures a negative host-indegree/quality correlation on ClueWeb22-A, and
WebGraphMix rejects centrality for post-hoc selection.
We reproduce the finding across ten classifiers on CC data and find that language differences partly explain the observed patterns.

A host's primary language covaries with its centrality, and the
English-trained quality classifiers score English and non-English hosts very
differently \citep{lucy2024aboutme}, but not in a single direction (Figure~\ref{fig:rq2b}). 
\DCLM is the extreme case since it rates non-English hosts far higher than English ones (median \valDCLMmedNonEng{}  %
vs.\ \valDCLMmedEng{}), so pooling the two languages makes its overall \HC correlation  %
\emph{negative} ($\rho = \valDCLMhcAll$) even though it is positive within English hosts  %
($\rho = \valDCLMhcEng$). Nemotron-Q leans the opposite way, scoring English hosts higher  %
(\valNemQmedEng{} vs.\ \valNemQmedNonEng{}). Multilingual classifiers, which do not share  %
the bias, already show positive signal on non-English hosts (JQL-Edu $\valJQLnonEng$,  %
FineWeb2-HQ $\valFWtwoNonEng$; full language-stratified table in  %
Appendix~\ref{app:corr_tables}). Centrality is thus entangled with language and not a clean quality signal.

\subsection{RQ3: A Small Per-Host Sample Decides Crawl Depth}
\label{sec:rq4}

\begin{figure}[htp]
\centering
\includegraphics[width=\columnwidth]{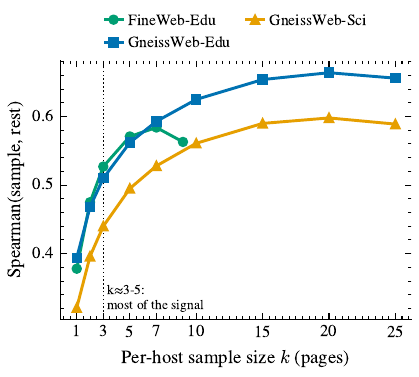}
\caption{
The correlation
between the quality of $k$ sampled pages and the quality of the host's remaining pages, where
$1.0$ is perfect prediction and $0$ is none. The curve climbs with $k$ and plateaus by
$k\approx15$--$20$, but a $\sim$3--5 page sample already captures most of the signal, so a
few pages suffice to decide crawl depth cheaply.
Shown for our \FWEdu sample (hosts with $\ge 10$ pages, so $k$ reaches $9$) and full-crawl GneissWeb ($\ge 30$ pages).
}
\label{fig:rq4a}
\end{figure}

\begin{figure}[htb]
\centering
\includegraphics[width=\columnwidth]{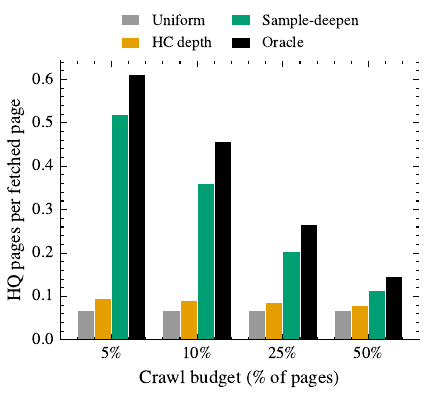}
\caption{Crawl efficiency based on \FWEdu, measured as high-quality pages obtained per page actually fetched, and for four
ways of spending a fixed budget across hosts. 
\emph{Sample-then-deepen} (our host-level crawl policy) nearly matches the \emph{oracle} and
collects several times more high-quality pages per fetch than crawling every host equally
(\emph{uniform breadth}) or biasing depth toward high-centrality hosts (\emph{\HC depth}).
}
\label{fig:rq4b}
\end{figure}

We now ask \emph{how deep} to crawl each host. Because quality is a host property (RQ1), a few pages sampled from a
host predict the rest: 
Figure~\ref{fig:rq4a} shows the rank correlation between a $k$-page sample and a host's
remaining pages rises with $k$ and plateaus by $k \approx 15$--$20$. The gain per added
page shrinks steadily, so a small $3$--$5$ page sample ($\rho = \valSampleRhoKthree{}$ at  %
$k{=}3$, $\valSampleRhoKfive{}$ at $k{=}5$) already carries most of the signal.  %
Ranking hosts on a three-page sample and deepening the best hosts already recovers most of the
oracle's depth yield. A larger sample would rank hosts better at more computational costs.

This motivates a \emph{sample-then-deepen} crawl policy, which samples and classifies a few pages per host,
then spends depth on the hosts whose sample are higher quality. Such a policy obtains up to
\valDeepenMultHi{}$\times$ more high-quality pages per fetched page than the uniform breadth policy  %
at small budgets (down to \valDeepenMultLo{}$\times$ as the budget grows) and approaches  %
the oracle policy. 

In contrast, depth biased by harmonic centrality (similar to what CC does
today) barely differs from uniform breadth (Figure~\ref{fig:rq4b}). 
The \emph{sample-then-deepen} policy has also the pratical advantage that the classifier is not in the crawler's hot path: unlike a page-level scorer (e.g., \CFLLM) it runs on a $\sim$3-page
sample per host, not on every page.

\subsection{RQ4: Common Crawl's Budget Is Blind to LLM Data Quality}
\label{sec:rq5}

\begin{figure}[tbp]
\centering
\includegraphics[width=\columnwidth]{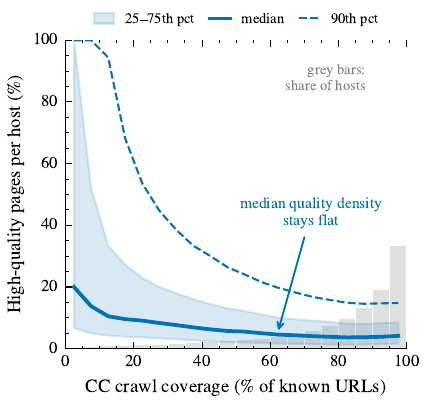}
\caption{Host-level quality density does not rise with crawl coverage (Spearman \valRhoLead{}).  %
The median quality density
stays flat and low across coverage (from \valMedQualLo{}\% at the least-crawled hosts to  %
\valMedQualHi{}\% at the near-fully-crawled majority). Grey bars show the  %
share of hosts in each coverage bin: most are already almost fully crawled, which is why the
overall rank correlation is near zero.}
\label{fig:rq5a}
\end{figure}

\begin{figure}[tbp]
\centering
\includegraphics[width=\columnwidth]{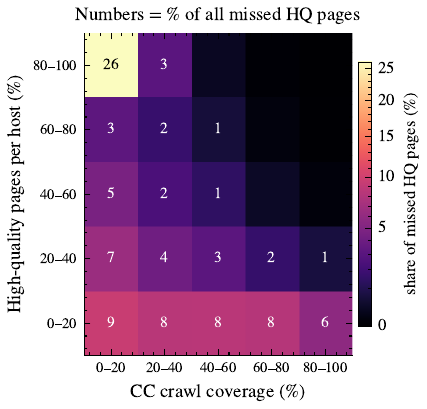}
\caption{The missed high-quality budget concentrates on barely-crawled hosts. Each cell gives
the share of the estimated $\approx$\,\valMissedBudgetM{}M uncrawled high-quality pages that  %
falls in one coverage$\times$quality bin (columns: crawl coverage; rows: quality density; axes
as in Figure~\ref{fig:rq5a}). Half (\valLowCovShare{}\%) of the missed pages lie on hosts  %
crawled below 20\%, and the single densest ``hidden-gem'' bin holds \valGemCellShare{}\% of missed high-quality pages alone.  %
}
\label{fig:rq5b}
\end{figure}

Finally, we turn from simulation to the deployed crawl. Using historical per-host
budget from CC-MAIN-2024-10 and GneissWeb quality scores, we find its crawl coverage carries no positive quality signal: across
\valLeadHosts{} hosts, the correlation between how fully a host is crawled and its quality  %
density is \valRhoLead{}, and the quality distribution does not rise with  %
coverage (Figure~\ref{fig:rq5a}). If anything the least-crawled hosts are the highest quality.
Assuming that uncrawled pages have the same quality density as
fetched pages from the same host, we estimate
\valMissedBudgetM{}M high-quality pages among the known URLs  %
left uncrawled in this one crawl. This missed budget concentrates on barely-crawled hosts:
\valLowCovShare{}\% of it lies on hosts crawled below 20\%, and  %
\valGemShare{}\% falls on a compact set of \valGemHosts{} ``hidden-gem'' hosts (below-median  %
coverage but top-decile quality) that represent only \valGemPctHosts{}\% of all hosts  %
(Figure~\ref{fig:rq5b}).
The URLs are known, but the quality of their unfetched pages
has not been directly measured.

Expressed as pretraining tokens, this single-crawl shortfall is on the order of tens of
billions of high-quality tokens (roughly \valMissedBudgetM{}\,billion at an assumed  %
$\sim$1{,}000 tokens per page). 
We find shortfalls of a similar scale in each crawl examined
(Table~\ref{tab:rq5_panel}). Extrapolating this estimate to
the +100 existing CC crawls would give approximately 6 trillion tokens before
accounting for repeated pages or duplicate content. This is
an illustrative extrapolation. We have not measured overlap
across crawls or established that the same shortfall applies
throughout the whole crawl archive.

\begin{table}[tbp]
\centering\footnotesize
\begin{tabular}{lrrr}
\toprule
Crawl & $\rho$(cov,\,qual) & Gems & Uncrawled HQ (M) \\
\midrule
2021-49 & $-0.151$ & 339{,}430 & 20.3 \\
2022-49 & $-0.057$ & 337{,}224 & 28.9 \\
2023-23 & $-0.060$ & 325{,}545 & 20.1 \\
2023-40 & $-0.079$ & 390{,}185 & 38.0 \\
2024-10 & $-0.081$ & 369{,}534 & 35.7 \\
2024-18 & $-0.061$ & 353{,}728 & 24.1 \\
\bottomrule
\end{tabular}

\caption{The coverage--quality disconnect is not a one-off. For each of \valGemCrawls{} crawls  %
(2021--2024) we report the Spearman correlation between crawl coverage and quality density
(all near $0$ and slightly negative, i.e.\ no positive alignment), the number of under-crawled
high-quality ``gem'' hosts, and the estimated high-quality URLs those hosts leave uncrawled.
Every crawl shows the same pattern and tens of millions of missed URLs.}
\label{tab:rq5_panel}
\end{table}

\begin{figure}[tbp]
\centering
\includegraphics[width=\columnwidth]{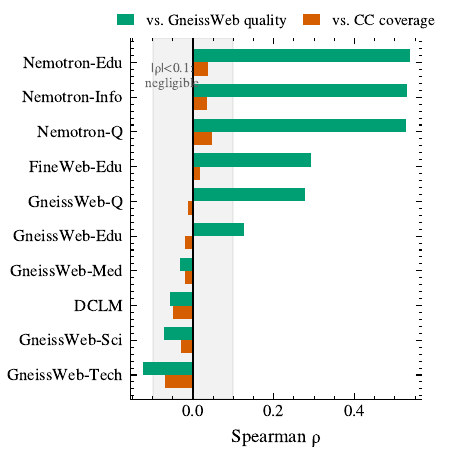}
\caption{Coverage-blindness does not depend on the GneissWeb filter. Per classifier, on one
Spearman axis: its agreement with the GneissWeb quality density (green) versus the correlation
of Common Crawl coverage with it (orange). Quality classifiers (Nemotron, \FWEdu) agree
substantially with GneissWeb (\DCLM and the topic scorers less so), yet every coverage bar
stays inside the negligible band ($|\rho|<0.1$): whichever quality signal we use, how fully a
host is crawled does not track how good it is.}
\label{fig:rq5rob}
\end{figure}

\paragraph{Stable and classifier-robust.} The pattern is not a one-crawl or one-classifier
artifact. Across six crawls spanning 2021--2024 the coverage--quality correlation stays
near zero and negative (\valGemRhoHi{} to \valGemRhoLo{}; Table~\ref{tab:rq5_panel}).  %
Also, the GneissWeb quality signal is confirmed by independent classifiers: Nemotron
(\valNemAgreeLo{}--\valNemAgreeHi{}) and \FWEdu ($\valFWeduAgree$) agree with it, while  %
Common Crawl coverage is uncorrelated with \emph{every} classifier we test
($|\rho| \le \valCovBlindMax{}$; Figure~\ref{fig:rq5rob}). \DCLM is the lone  %
non-agreeing classifier ($\valDCLMagree$), consistent with its known orthogonality to  %
other quality signals (\S\ref{sec:rq2}). The missed budget skews toward national and
regional news publishers and scholarly repositories: the \texttt{.uk} TLD alone accounts
for \valUKbudget{}\% of it from \valUKhosts{}\% of hosts, at just \valUKcov{}\% median  %
coverage (Appendix, Figure~\ref{fig:rq5char}). %

\section{Discussion}
\label{sec:discussion}

\paragraph{A host-level companion to \CFLLM.} \CFLLM \citep{yu2025craw4llm} prioritizes the crawl frontier by
the quality score of each fetched page. Thus, a crawler must fetch, extract, and score every
candidate to obtain that signal. Our results address a different, complementary lever.
Because quality is a host property (RQ1) and a $\sim$3-page sample predicts a host's rest (RQ3), the scheduling signal for \emph{how deep} to crawl a host can be obtained from a small sample rather than from every page, and the classifier need not live in the crawler's hot path. 
We answer ``how much to crawl each host, and which under-crawled
hosts are worth more,'' which the page-level, host-blind formulation does not address. 
Our analysis complements page-level scheduling and informs how much crawl budget should be allocate to each host.

\paragraph{Explaining earlier findings on graph centrality.} \CFLLM and WebGraphMix \citep{badoni2026webgraphmix} report
limited benefits from the centrality measures they evaluate.
Our crawl-yield results show a similar pattern (RQ2):
centrality-based ordering performs much worse than the oracle
and, for \DCLM, worse than random ordering.

Language differences help explain some of the observed
centrality--quality correlations. English-trained classifiers
score English and non-English hosts differently. For instance,
\DCLM assigns higher scores to non-English hosts. Centrality
also varies by language. The correlation between \HC and
\DCLM scores is negative when all hosts are pooled, but positive
when only English-language hosts are considered.
High-centrality hosts also contain a mix of high- and low-quality
pages, limiting centrality's usefulness as a quality signal.

However, the language analysis does not establish that accounting
for language improves crawl yield. A positive correlation within
a language group need not translate into an effective ranking
under a limited crawl budget. Likewise, positive correlations
with multilingual classifiers do not by themselves demonstrate
a better crawl policy.
Lastly, centrality could still be used to group hosts for sampling,
although we do not evaluate whether this improves crawl yield.

\paragraph{What the deployed crawl leaves on the table.} RQ4 turns the diagnostic on CC's deployed crawl policy itself. Its per-host budget is uncorrelated with quality under every classifier we tested, so the crawl is not preferentially deepening the hosts worth deepening. 
This misalignment is expected rather than a defect: Common Crawl was built as a broad, general-purpose sample of the Web serving diverse use cases, not as an LLM-oriented crawler \citep{baack2024critical}.

Assuming that uncrawled pages have the same quality density as fetched pages from the same host, we estimate that tens of millions of high-quality pages remain among known but uncrawled URLs in each crawl.
The presented \emph{sample-then-deepen} approach could help recover these pages by allocating more budget to hosts with high-scoring samples.
It is worth noting that this gap is about \textit{known} URLs, for an estimation of whole URL population we refer to \citet{paris2026measuringcrawlerseesdiscovery}.

\paragraph{Quality and discovery are different objectives.}
Our depth simulations allocate a crawl budget over a fixed set of
already-fetched pages. They do not measure which additional
URLs or hosts a crawler would discover by following links.
A host with low classifier scores could nevertheless provide
links to useful content elsewhere. Weak performance as a
quality-ranking signal therefore does not rule out a role
for centrality in discovery. A crawler could combine
centrality-based exploration with sample-guided depth
allocation, but our experiments do not evaluate this
combination.

\section{Conclusions}
\label{sec:conclusions}

In this paper, we studied Web crawl policies for LLM pretraining data at the level of hosts and reached
the following conclusions.

Document quality is largely a host property, i.e, between-host variance is about half of the
total. This finding motivates the host-aware policy and
lets a small sample stand in for a host.
Ordering hosts by graph centrality (\HC or \PR) collects at most
\valYieldHChi{} times as many high-quality pages as random ordering,  %
and fewer for \DCLM. An oracle policy that knows each host's quality density
collects \valYieldOracleLo{}--\valYieldOracleHi{} times as many.  %
Centrality is a weak guide partly because it varies with host language,
which also affects classifier scores, and because highly central hosts
contain a mix of high- and low-quality pages.

In our crawl simulations, a three-page sample provides a useful signal for allocating crawl depth. The sample-then-deepen policy approaches the oracle high-quality yield, reaching up to \valDeepenMultHi{} times the yield of the uniform breadth policy, whereas centrality-biased depth provides little to none improvement.  %
Finally, Common Crawl's deployed crawl budget is itself blind to LLM data quality. The crawl coverage is uncorrelated
with quality (Spearman \valRhoLead{}) under every classifier, leaving an estimated  %
\valMissedBudgetM{}M known high-quality URLs uncrawled in a single crawl, \valGemShare{}\% of them on  %
a compact set of \valGemHosts{} under-crawled hosts.  %
This finding can be attributed to Common Crawl aiming for a broad, general-purpose sample of Web, and not LLM pretraining data.

In summary, our results motivate a crawl policy that first samples a few pages
from each host, scores their quality, and then fetches more pages
from hosts with higher-scoring samples. This could recover
high-quality pages missed by the current crawl policy.
Graph centrality could help group hosts for sampling, but should
not be used on its own to judge their content quality.

\subsection{Future Work}
\label{sec:future}

\textbf{A live sample-then-deepen crawl.} All four
research questions rely on existing data from Common Crawl
and did not involve any live crawling. RQ1 measures host-level variation in classifier
scores, while RQ4 examines observed crawl coverage and quality
density. RQ2 and RQ3 additionally simulate alternative crawl
allocations over already-fetched pages. None of these analyses
tests a newly deployed crawler under production settings, future work needs to address this.
In particular, a live crawl experiment must account for fetch latency, content freshness, politeness limits, and the discovery of new URLs and hosts.

\textbf{Recovering the hidden gems.} RQ4 identifies tens of millions of known
high-quality URLs on under-crawled hosts. A targeted re-crawl of the top gem hosts,
followed by a downstream ablation, would quantify how much the missed budget is worth in
trained-model terms.

\textbf{Downstream model ablation.} We measure crawl yield and coverage, not trained-model
performance. Given that proxy metrics and downstream accuracy can diverge
\citep{li2024datacomp,yu2025craw4llm}, one would need to train small models ($\sim$1B parameters) on data
selected by uniform, centrality, and sample-then-deepen policies test whether
the yield gains translate into downstream quality.

\textbf{Beyond a single quality filter.} The quality-density signal uses primarily a single
filter (GneissWeb) due to the availability of precomputed annotations. Combining several classifiers, and modeling their per-host agreement,
may sharpen both the depth decision and the hidden-gem ranking.

\section*{Limitations}

The following discussion complements the limitations described in \S\ref{sec:future}.

\textbf{Quality density is a proxy.} Our per-host quality signal is the fraction of a
host's crawled pages retained by a classifier model (GneissWeb). It inherits that models's
biases. We mitigate this by showing the RQ4 result holds under independent models
(Nemotron, \FWEdu), but no model is ground truth \citep{gururangan2022whose}.

\textbf{Missed-budget estimate.} The uncrawled-high-quality-URL count multiplies a host's
uncrawled known URLs by its observed quality density, assuming the uncrawled pages resemble
the crawled ones. It is an estimate of scale, not an exact count. 
Furthermore, the estimate does not account for overlap between crawls and other factors like crawl politeness or content freshness.

\textbf{Selection of pages and hosts.} Both the host-consistency
estimates and the sample-size analysis concern pages already
selected by Common Crawl. Our WARC sample introduces a further
sampling step, and minimum-document thresholds exclude sparsely
represented hosts. The full-crawl GneissWeb analysis removes the
WARC subsampling step, but still reflects Common Crawl's page
selection. These choices may affect measured within-host
variation and how well a small sample predicts other pages.
Our results therefore do not establish that three pages suffice
for arbitrary hosts or for pages outside the observed crawls.

\textbf{Representation and diversity.} Allocating more crawl
budget to hosts with high classifier scores may reduce the
representation of languages, regions, or content types that
those classifiers score poorly. At crawl time, this affects
which pages are collected at all, including potentially useful
pages on hosts whose initial samples receive low scores.
Higher classifier-based yield therefore does not necessarily
imply a more representative or diverse corpus. A deployment
could reserve budget for broader exploration and monitor
coverage across languages and content types, but the
effectiveness of these measures remains to be evaluated.

\bibliography{references}

\appendix

\setcounter{topnumber}{2}
\setcounter{bottomnumber}{1}
\setcounter{totalnumber}{3}
\renewcommand{\topfraction}{0.7}
\renewcommand{\bottomfraction}{0.3}
\renewcommand{\textfraction}{0.2}
\renewcommand{\floatpagefraction}{0.5}
\renewcommand{\dbltopfraction}{0.9}     %
\renewcommand{\dblfloatpagefraction}{0.5}

\newpage

\section{Temporal Stability of Graph Ranks}
\label{app:rank_stability}

\begin{table}[h]
\centering\footnotesize
\resizebox{\columnwidth}{!}{%
\begin{tabular}{llrrr}
\toprule
Later release & Gap & $N$ hosts & \HC $\rho$ & \PR $\rho$ \\
\midrule
2024-may-jun-jul & $\sim$3-5mo after & 531{,}054 & $0.875$ & $0.923$ \\
2024-25-nov-dec-jan & $\sim$9-11mo after & 527{,}432 & $0.870$ & $0.900$ \\
2025-may-jun-jul & $\sim$15mo after & 524{,}077 & $0.851$ & $0.885$ \\
\bottomrule
\end{tabular}
}
\caption{Spearman rank correlation between a host's centrality rank in our reference graph
and its rank in later Common Crawl graph releases, where $1.0$ would mean an identical
ordering of all hosts. Both harmonic centrality (\HC) and PageRank (\PR) stay above $0.85$
even $\sim$15 months apart, so host rankings barely move over a year. This stability is what
lets us pair one aligned graph with document snapshots from different crawls instead of
rebuilding the graph each time.}
\label{tab:rq3_stability}
\end{table}

We assess how host centrality rankings change over time by comparing
the reference graph with releases up to approximately 15 months later.
Host-level \HC and \PR ranks have Spearman correlations of
\valHCstabLo{}--\valPRstabHi{} across these comparisons  %
(Table~\ref{tab:rq3_stability}). This stability supports using
a single reference graph for the document snapshots in our study.

\section{Crawl-Yield Detail (RQ2)}
\label{app:rq2_yield}

\begin{table}[h]
\centering\footnotesize
\resizebox{\columnwidth}{!}{%
\begin{tabular}{lrrrrr}
\toprule
Classifier & Current & \HC & \PR & Oracle & TLDs (HC/rnd) \\
\midrule
\DCLM & $0.75\times$ & $0.67\times$ & $0.73\times$ & $\mathbf{3.98\times}$ & 223\,/\,422 \\
\FWEdu & $1.60\times$ & $1.58\times$ & $1.47\times$ & $\mathbf{7.97\times}$ & 223\,/\,422 \\
Nemotron Quality & $1.27\times$ & $1.79\times$ & $1.56\times$ & $\mathbf{4.05\times}$ & 223\,/\,422 \\
\bottomrule
\end{tabular}
}
\caption{High-quality documents collected at a 10\% crawl budget, as a multiple of a random
crawl ($1.0=$random, $2\times=$twice as many). Centrality ordering (\HC, \PR) gains at most
\valYieldHChi{}$\times$---and drops \emph{below} $1$ for \DCLM---while the oracle reaches  %
\valYieldOracleLo{}--\valYieldOracleHi{}$\times$, so most of the achievable gain is left  %
untaken. The last column shows centrality ordering also roughly halves the number of
top-level domains represented (HC vs.\ random), narrowing corpus diversity. This is the
per-classifier, fixed-budget companion to Figure~\ref{fig:rq2a}; the full budget sweep is
Figure~\ref{fig:rq2_curves}.}
\label{tab:rq2_yield}
\end{table}

This section gives the per-classifier crawl-yield numbers behind \S\ref{sec:rq2}.
Table~\ref{tab:rq2_yield} fixes the budget at 10\% of pages and reports, for each classifier,
the high-quality yield of each ordering policy as a multiple of a random crawl, together with
the TLD diversity the policy retains; Figure~\ref{fig:rq2_curves} plots the same comparison
across all budgets. It makes the two headline points of RQ2 concrete at a single budget:
centrality ordering barely lifts the yield above random (and drops below it for \DCLM) while
the oracle is several times higher, and centrality ordering simultaneously narrows the range of
top-level domains crawled.

\section{Full Graph-Classifier Correlation Tables}
\label{app:corr_tables}

These tables expand the correlation analysis behind RQ1--RQ2 using alternative classifiers, host-inclusion thresholds, and aggregation methods.

\subsection{HC and PageRank Correlations Across Classifiers}

Table~\ref{tab:app_graph_clf} reports correlations of host-level
\HC and \PR with all twelve classifiers, using Spearman $\rho$,
Kendall $\tau$, and weighted Spearman $\rho$. Spearman correlations
range from $-0.086$ to $+0.207$. For each classifier, \HC and \PR
have the same correlation sign and similar magnitudes.

\begin{table}[ht]
\centering\footnotesize
\resizebox{\columnwidth}{!}{%
\begin{tabular}{llrrrr}
\toprule
Metric & Classifier & Spearman $\rho$ & Kendall $\tau$ & Weighted $\rho$ & $N$ hosts \\
\midrule
$\log\HC$ & \DCLM & $-0.083$ & $-0.055$ & $-0.087$ & 27{,}910 \\
$\log\HC$ & GneissWeb Quality & $+0.021$ & $+0.015$ & $+0.018$ & 27{,}910 \\
$\log\HC$ & GneissWeb Sci & $+0.031$ & $+0.021$ & $+0.030$ & 27{,}910 \\
$\log\HC$ & GneissWeb Edu & $+0.039$ & $+0.026$ & $+0.037$ & 27{,}910 \\
$\log\HC$ & GneissWeb Tech & $+0.036$ & $+0.024$ & $+0.038$ & 27{,}910 \\
$\log\HC$ & GneissWeb Med & $-0.032$ & $-0.021$ & $-0.035$ & 27{,}910 \\
$\log\HC$ & \FWEdu & $+0.122$ & $+0.083$ & $+0.125$ & 27{,}910 \\
$\log\HC$ & Nemotron Quality & $+0.167$ & $+0.110$ & $+0.169$ & 27{,}910 \\
$\log\HC$ & Nemotron Edu & $+0.156$ & $+0.106$ & $+0.154$ & 27{,}910 \\
$\log\HC$ & Nemotron Info & $+0.149$ & $+0.100$ & $+0.151$ & 27{,}910 \\
$\log\HC$ & JQL-Edu & $+0.124$ & $+0.083$ & $+0.120$ & 15{,}661 \\
$\log\HC$ & FineWeb2-HQ & $+0.207$ & $+0.140$ & $+0.217$ & 13{,}614 \\
$\log\PR$ & \DCLM & $-0.086$ & $-0.057$ & $-0.089$ & 27{,}910 \\
$\log\PR$ & GneissWeb Quality & $+0.012$ & $+0.008$ & $+0.009$ & 27{,}910 \\
$\log\PR$ & GneissWeb Sci & $+0.015$ & $+0.010$ & $+0.015$ & 27{,}910 \\
$\log\PR$ & GneissWeb Edu & $+0.023$ & $+0.015$ & $+0.021$ & 27{,}910 \\
$\log\PR$ & GneissWeb Tech & $+0.029$ & $+0.019$ & $+0.032$ & 27{,}910 \\
$\log\PR$ & GneissWeb Med & $-0.039$ & $-0.026$ & $-0.042$ & 27{,}910 \\
$\log\PR$ & \FWEdu & $+0.110$ & $+0.075$ & $+0.113$ & 27{,}910 \\
$\log\PR$ & Nemotron Quality & $+0.170$ & $+0.113$ & $+0.172$ & 27{,}910 \\
$\log\PR$ & Nemotron Edu & $+0.148$ & $+0.101$ & $+0.147$ & 27{,}910 \\
$\log\PR$ & Nemotron Info & $+0.152$ & $+0.101$ & $+0.153$ & 27{,}910 \\
$\log\PR$ & JQL-Edu & $+0.079$ & $+0.053$ & $+0.076$ & 15{,}661 \\
$\log\PR$ & FineWeb2-HQ & $+0.173$ & $+0.117$ & $+0.184$ & 13{,}614 \\
\bottomrule
\end{tabular}
}
\caption{Host-level correlations of harmonic centrality and
  PageRank with twelve classifier scores. Spearman $\rho$,
  Kendall $\tau$, and weighted Spearman $\rho$ agree in sign
  for every metric--classifier pair.}
\label{tab:app_graph_clf}
\end{table}

\subsection{Language-Stratified Correlations ($\log\HC$, $n \geq 3$)}

Table~\ref{tab:app_lang_strat} is the full version of the language split behind
Figure~\ref{fig:rq2b}: the \HC--quality correlation for every classifier, computed on all
hosts and separately on English- and non-English-primary hosts. It shows the sign flip is
general (classifiers tend to correlate more positively within a single language than pooled
across languages) and that the multilingual classifiers (JQL-Edu, FineWeb2-HQ) are the ones
that retain a positive signal on non-English hosts.

\begin{table}[ht]
\centering\footnotesize
\begin{tabular}{llrr}
\toprule
Stratum & Classifier & Spearman $\rho$ & $N$ hosts \\
\midrule
English & \FWEdu & $+0.237$ & 13{,}670 \\
English & FineWeb2-HQ & $+0.233$ & 1{,}109 \\
English & GneissWeb Sci & $+0.199$ & 13{,}670 \\
English & Nemotron Edu & $+0.157$ & 13{,}670 \\
English & GneissWeb Tech & $+0.154$ & 13{,}670 \\
English & GneissWeb Edu & $+0.143$ & 13{,}670 \\
English & Nemotron Quality & $+0.127$ & 13{,}670 \\
English & GneissWeb Med & $+0.111$ & 13{,}670 \\
English & Nemotron Info & $+0.097$ & 13{,}670 \\
English & JQL-Edu & $+0.083$ & 1{,}421 \\
English & GneissWeb Quality & $+0.072$ & 13{,}670 \\
English & \DCLM & $+0.064$ & 13{,}670 \\
Non-English & FineWeb2-HQ & $+0.186$ & 12{,}505 \\
Non-English & JQL-Edu & $+0.137$ & 14{,}186 \\
Non-English & Nemotron Edu & $+0.043$ & 14{,}186 \\
Non-English & \FWEdu & $+0.023$ & 14{,}186 \\
Non-English & GneissWeb Sci & $+0.023$ & 14{,}186 \\
Non-English & GneissWeb Quality & $+0.017$ & 14{,}186 \\
Non-English & \DCLM & $+0.017$ & 14{,}186 \\
Non-English & Nemotron Info & $+0.016$ & 14{,}186 \\
Non-English & Nemotron Quality & $+0.012$ & 14{,}186 \\
Non-English & GneissWeb Edu & $+0.002$ & 14{,}186 \\
Non-English & GneissWeb Med & $+0.001$ & 14{,}186 \\
Non-English & GneissWeb Tech & $-0.007$ & 14{,}186 \\
\bottomrule
\end{tabular}

\caption{Full language-stratified \HC correlations. For JQL-Edu and FineWeb2-HQ,
``English'' rows reflect host-primary language classification; inference excluded
documents mapped as English.}
\label{tab:app_lang_strat}
\end{table}

\subsection{TLD Macro-Category Breakdown}

Table~\ref{tab:app_tld_category} breaks the \HC--\DCLM correlation down by TLD macro-category.
It is negative precisely within the academic and commercial categories, which concentrate the
non-English hosts that \DCLM scores highly, i.e., localising the language confound of
\S\ref{sec:rq2} to the domains where language and centrality covary most.

\begin{table}[H]
\centering\footnotesize
\begin{tabular}{lrr}
\toprule
TLD Category & HC vs.\ \DCLM $\rho$ & $N$ hosts \\
\midrule
Other & $-0.159$ & 659 \\
Academic (.edu, .ac.*) & $-0.098$ & 766 \\
Commercial (.com, .net) & $-0.028$ & 13{,}470 \\
Country-code & $+0.004$ & 10{,}939 \\
Nonprofit (.org) & $+0.032$ & 1{,}881 \\
Government (.gov.*) & $+0.050$ & 195 \\
\bottomrule
\end{tabular}

\caption{HC--\DCLM Spearman $\rho$ by TLD macro-category. Negative within academic
and commercial TLDs reflects DCLM's language inflation of non-English domains.}
\label{tab:app_tld_category}
\end{table}

\subsection{Score Distributions by Language Stratum}

Table~\ref{tab:app_score_dist} gives the full score distributions (mean, median, and
percentiles) behind Figure~\ref{fig:rq2b}, split by host language. 
\DCLM's median score is substantially higher on non-English hosts than on English hosts, whereas \FWEdu assigns similar scores to the two groups. 
These differences are consistent with language composition contributing to the pooled centrality-quality correlation.

\begin{table}[H]
\centering\footnotesize
\resizebox{\columnwidth}{!}{%
\begin{tabular}{llrrrr}
\toprule
Stratum & Classifier & Mean & Median & P10 & P90 \\
\midrule
English & \DCLM & 0.041 & 0.001 & 0.000 & 0.065 \\
English & \FWEdu & 0.220 & 0.203 & 0.041 & 0.408 \\
English & Nemotron Quality & 0.541 & 0.487 & 0.396 & 0.785 \\
English & JQL-Edu & -4.850 & -5.409 & -8.946 & 0.190 \\
English & FineWeb2-HQ & 0.242 & 0.029 & 0.000 & 0.958 \\
Non-English & \DCLM & 0.367 & 0.289 & 0.002 & 0.892 \\
Non-English & \FWEdu & 0.223 & 0.211 & 0.020 & 0.417 \\
Non-English & Nemotron Quality & 0.352 & 0.369 & 0.238 & 0.400 \\
Non-English & JQL-Edu & -4.478 & -4.897 & -8.167 & -0.096 \\
Non-English & FineWeb2-HQ & 0.132 & 0.005 & 0.000 & 0.586 \\
\bottomrule
\end{tabular}
}
\caption{Score distributions by language stratum (selected classifiers).
JQL-Edu scores are log-probabilities.}
\label{tab:app_score_dist}
\end{table}

\subsection{Full ICC1 Table}

Table~\ref{tab:app_full_icc} lists the full RQ1 variance decomposition: ICC1 for
every classifier at $n \geq 3$ documents per host. The between-host share is
stable across the $n \geq 3$, $5$, and $10$ thresholds and near one-half throughout, the
property that lets a small per-host sample stand in for a host (\S\ref{sec:rq1}).

\begin{table}[H]
\centering\footnotesize
\resizebox{\columnwidth}{!}{%
\begin{tabular}{lrrrrrr}
\toprule
Classifier & $n_\text{hosts}$ & $n_\text{docs}$ & $\bar m$ & $\sigma^2_\text{btw}$ & $\sigma^2_\text{wtn}$ & ICC1 \\
\midrule
Nemotron Quality & 27{,}950 & 104{,}098 & 3.72 & 0.0195 & 0.0105 & 0.650 \\
Nemotron Info & 27{,}950 & 104{,}098 & 3.72 & 0.0379 & 0.0222 & 0.630 \\
Nemotron Edu & 27{,}950 & 104{,}098 & 3.72 & 0.0281 & 0.0190 & 0.596 \\
GneissWeb Tech & 27{,}950 & 104{,}098 & 3.72 & 0.0634 & 0.0461 & 0.579 \\
\DCLM & 27{,}950 & 104{,}098 & 3.72 & 0.0596 & 0.0461 & 0.564 \\
\FWEdu & 27{,}950 & 104{,}098 & 3.72 & 0.0131 & 0.0109 & 0.545 \\
JQL-Edu & 15{,}690 & 52{,}687 & 3.36 & 7.3071 & 6.2660 & 0.538 \\
GneissWeb Med & 27{,}950 & 104{,}098 & 3.72 & 0.0307 & 0.0275 & 0.528 \\
GneissWeb Sci & 27{,}950 & 104{,}098 & 3.72 & 0.0338 & 0.0319 & 0.514 \\
FineWeb2-HQ & 13{,}639 & 45{,}320 & 3.32 & 0.0485 & 0.0495 & 0.495 \\
GneissWeb Edu & 27{,}950 & 104{,}098 & 3.72 & 0.0153 & 0.0176 & 0.465 \\
GneissWeb Quality & 27{,}950 & 104{,}098 & 3.72 & 0.0154 & 0.0181 & 0.459 \\
\bottomrule
\end{tabular}
}
\caption{Full ICC1 table ($n \geq 3$ docs per host). ICC1 stable across $n \geq 3/5/10$ thresholds.}
\label{tab:app_full_icc}
\end{table}

\section{Datasets and Models}
\label{app:datasets}

Our study combines several public Common Crawl artifacts and Hugging Face models across the
two data regimes of \S\ref{sec:methods}. Table~\ref{tab:datasets} lists every dataset and
which research question uses it. Table~\ref{tab:models} lists the exact classifier models we
run in the classified-sample regime.

\begin{table*}[]
\centering\footnotesize
\renewcommand{\arraystretch}{1.5}
\begin{tabular}{p{3.5cm}p{9.6cm}}
\toprule
Dataset & Description and role (research question) \\
\midrule
\href{https://data.commoncrawl.org/crawl-data/CC-MAIN-2024-10/index.html}{CC-MAIN-2024-10} & Crawl (WARC) for primary crawl; we extract text (Trafilatura) and score it ourselves (224 WARCs, RQ1--RQ4). \\
\href{https://data.commoncrawl.org/crawl-data/CC-MAIN-2023-40/index.html}{CC-MAIN-2023-40} & Crawl (WARC) for second snapshot and the language/robustness checks (32 WARCs, RQ2). \\
CC-MAIN-\{2021-49, 2022-49, 2023-23, 2023-40, 2024-10, 2024-18\} & Crawl (host stats from six crawls) for the temporal panel (RQ4). \\
\href{https://index.commoncrawl.org/web-graphs-index.html}{Graph release (Sep 2023--Feb 2024)} & Aligned host-level graph with harmonic centrality and PageRank per host (RQ2, App.~\ref{app:rank_stability}). \\
Later graph releases (2024-May, 2024-Nov, 2025-May) & Host \HC/\PR ranks at $\sim$3--15 months later, for rank stability (App.~\ref{app:rank_stability}). \\
\href{https://huggingface.co/datasets/commoncrawl/host-index-testing-v2}{CC host index} & Per-host known/fetched URL counts and crawl coverage (RQ4). \\
\href{https://huggingface.co/datasets/commoncrawl/gneissweb-annotation-host-testing-v1}{GneissWeb annotations} & Precomputed annotations for full-crawl page-level quality labels (every fetched page); source of the per-host quality density (RQ3, RQ4). \\
\bottomrule
\end{tabular}
\caption{Datasets used, by data regime; the research question each supports is noted in
parentheses. WARC crawls are extracted and scored by us (classified-sample regime); the host
index and GneissWeb annotations are Common Crawl's precomputed full-crawl artifacts.}
\label{tab:datasets}
\end{table*}

\begin{table*}[]
\centering\footnotesize
\renewcommand{\arraystretch}{1.5}
\begin{tabular}{p{2.4cm}p{1.8cm}p{6.9cm}p{2.9cm}}
\toprule
Classifier & Family & Hugging Face model & Notes \\
\midrule
\DCLM & fastText & \texttt{mlfoundations/}\allowbreak\texttt{fasttext-oh-eli5} & OH-2.5\,+\,ELI5 vs.\ RefinedWeb, bigram \\
GneissWeb Quality & fastText & \texttt{ibm-granite/}\allowbreak\texttt{GneissWeb.Quality\_annotator} & quality annotator \\
GneissWeb Sci & fastText & \texttt{ibm-granite/}\allowbreak\texttt{GneissWeb.Sci\_classifier} & science topic \\
GneissWeb Edu & fastText & \texttt{ibm-granite/}\allowbreak\texttt{GneissWeb.Edu\_classifier} & education topic \\
GneissWeb Tech & fastText & \texttt{ibm-granite/}\allowbreak\texttt{GneissWeb.Tech\_classifier} & technology topic \\
GneissWeb Med & fastText & \texttt{ibm-granite/}\allowbreak\texttt{GneissWeb.Med\_classifier} & medical topic \\
\FWEdu & fastText & \texttt{kenhktsui/}\allowbreak\texttt{fineweb-edu-fasttext-classifier} & educational value; fastText re-impl. \\
Nemotron (Q/Edu/Info) & fastText & \texttt{nvidia/}\allowbreak\texttt{nemotron-climb-fasttext-classifiers} & three heads: quality, edu, info \\
JQL-Edu & Transformer & \texttt{JQL-AI/}\allowbreak\texttt{JQL-Edu-Heads} on \texttt{Snowflake/}\allowbreak\texttt{snowflake-arctic-embed-m-v2.0} & multilingual (35 lang.); non-English only \\
FineWeb2-HQ & Transformer & \texttt{epfml/}\allowbreak\texttt{FineWeb-HQ-Classifiers} on \texttt{FacebookAI/}\allowbreak\texttt{xlm-roberta-base} & multilingual (21 lang.); non-English only \\
\bottomrule
\end{tabular}
\caption{Exact Hugging Face models used to classify pages in the classified-sample regime.
\FWEdu is a compute-efficient fastText re-implementation \citep{ktsui2024fineweb} of the
original Transformer classifier, at comparable accuracy. The Transformer classifiers
(JQL-Edu, FineWeb2-HQ; the latter is the classifier of \citealp{messmer2025multilingual})
are frozen-embedding models with lightweight per-language heads and are run on non-English
documents only.}
\label{tab:models}
\end{table*}

\newpage
\newpage

\section{Supplementary Figures}
\label{app:figures}

These figures are the visual companions to results summarised in the body: the full
crawl-yield budget sweep behind Figure~\ref{fig:rq2a} (\S\ref{sec:rq2}), the temporal
robustness of the coverage--quality disconnect (Table~\ref{tab:rq5_panel}), the
composition of the missed high-quality budget (\S\ref{sec:rq5}), and the agreement structure
among the twelve classifiers (\S\ref{sec:rq2}).

\begin{figure*}[t]
\centering
\includegraphics[width=0.98\textwidth]{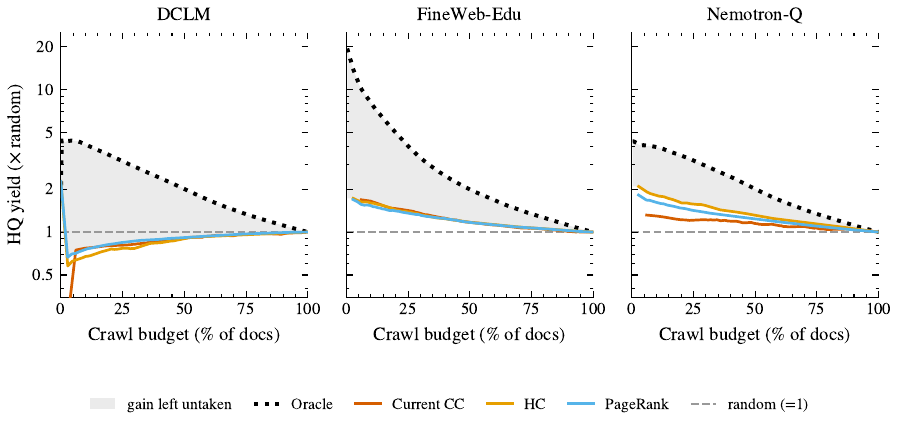}
\caption{Full budget sweep behind Figure~\ref{fig:rq2a}: high-quality documents collected as
the crawl budget grows, divided by what a random crawl collects at the same budget (log-scaled
$y$-axis; the dashed line at $1$ means ``no better than random''). Ordering hosts by harmonic
centrality (\HC), PageRank (\PR), or Common Crawl's current allocation hugs the random line and
stays far below the \emph{oracle}; the shaded band shows the gap between centrality-based yield and the oracle upper bound, e.g, widest at small budgets, where crawl order matters most, and closing as the budget
approaches a full crawl. For \DCLM, centrality is even worse than random.}
\label{fig:rq2_curves}
\end{figure*}

\paragraph{Crawl yield across budgets.} Figure~\ref{fig:rq2_curves} is the full-budget
companion to the fixed-budget bars of Figure~\ref{fig:rq2a}. Sweeping the crawl budget from a
small fraction of pages up to the full crawl shows that centrality ordering trails the oracle
at \emph{every} budget: the gap is widest at the small budgets where crawl order matters most
and closes only as the crawl approaches completion, so the fixed-10\% snapshot in the body is
representative rather than a cherry-picked operating point.

\begin{figure*}[t]
\centering
\includegraphics[width=0.98\textwidth]{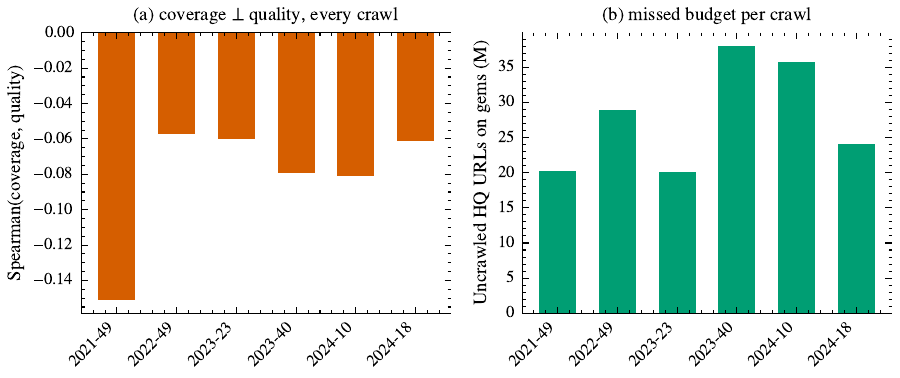}
\caption{The coverage--quality disconnect across \valGemCrawls{} Common Crawl snapshots (2021--2024).  %
(a) The Spearman correlation between crawl coverage and host quality density is near zero and
negative in every crawl (\valGemRhoHi{} to \valGemRhoLo{}). (b) Each crawl leaves tens of millions  %
of high-quality URLs uncrawled on hidden-gem hosts. The misallocation is a persistent, not a
single-crawl, artifact---the plotted form of Table~\ref{tab:rq5_panel}.}
\label{fig:rq5panel}
\end{figure*}

\paragraph{Persistence across crawls.} Figure~\ref{fig:rq5panel} is the plotted form of
Table~\ref{tab:rq5_panel}: the coverage--quality disconnect of RQ4 holds in each of
\valGemCrawls{} crawls spanning 2021--2024, both as a near-zero, slightly negative correlation  %
(panel a) and as tens of millions of uncrawled high-quality URLs per crawl (panel b). The
misallocation is thus a standing property of the deployed crawl, not a single-crawl artifact.

\begin{figure*}[t]
\centering
\includegraphics[width=0.98\textwidth]{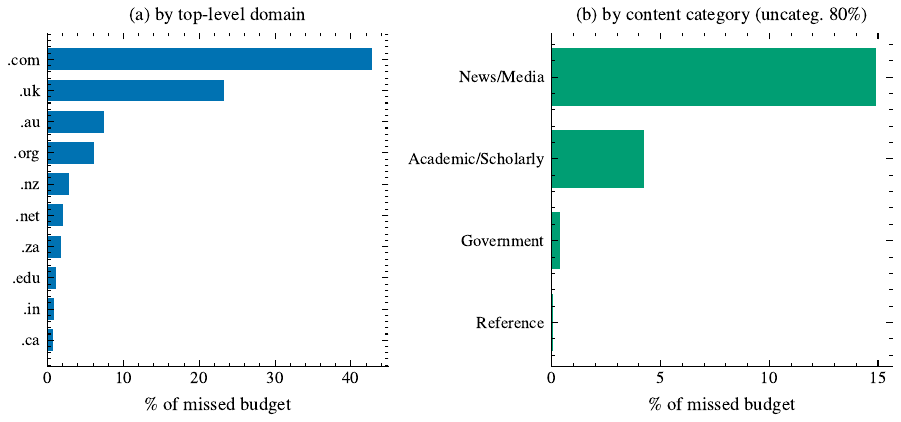}
\caption{Where the missed high-quality budget sits, by (a) top-level domain and (b) content
category (aggregates only; no host is named). Country-code publisher TLDs are over-represented
relative to their host count---\texttt{.uk} alone holds \valUKbudget{}\% of the missed budget from  %
just \valUKhosts{}\% of hosts at \valUKcov{}\% median coverage. Among the hosts a keyword classifier  %
can label, news/media and scholarly hosts dominate; the remaining $\sim$80\% is uncategorised (a
keyword lower bound), so panel (b) shows only the labelled share and omits the ``Other'' bucket.
``Reference'' denotes encyclopedic/reference sites (e.g.\ Wikipedia-family, dictionaries,
encyclopedias); it is near zero because the major reference sites are already well crawled, so
only a few dozen such hosts remain under-crawled gems.}
\label{fig:rq5char}
\end{figure*}

\paragraph{Where the missed budget sits.} Figure~\ref{fig:rq5char} characterises the
hidden-gem hosts in aggregate (no individual host is named). Country-code publisher TLDs---led
by \texttt{.uk}---are heavily over-represented relative to their host count, and among the
hosts a keyword classifier can label, news/media outlets and scholarly repositories dominate.
This is what the recoverable pool of \S\ref{sec:rq5} concretely consists of, and it motivates
the targeted re-crawl proposed in the future work.

\begin{figure*}[t]
\centering
\includegraphics[width=0.98\textwidth]{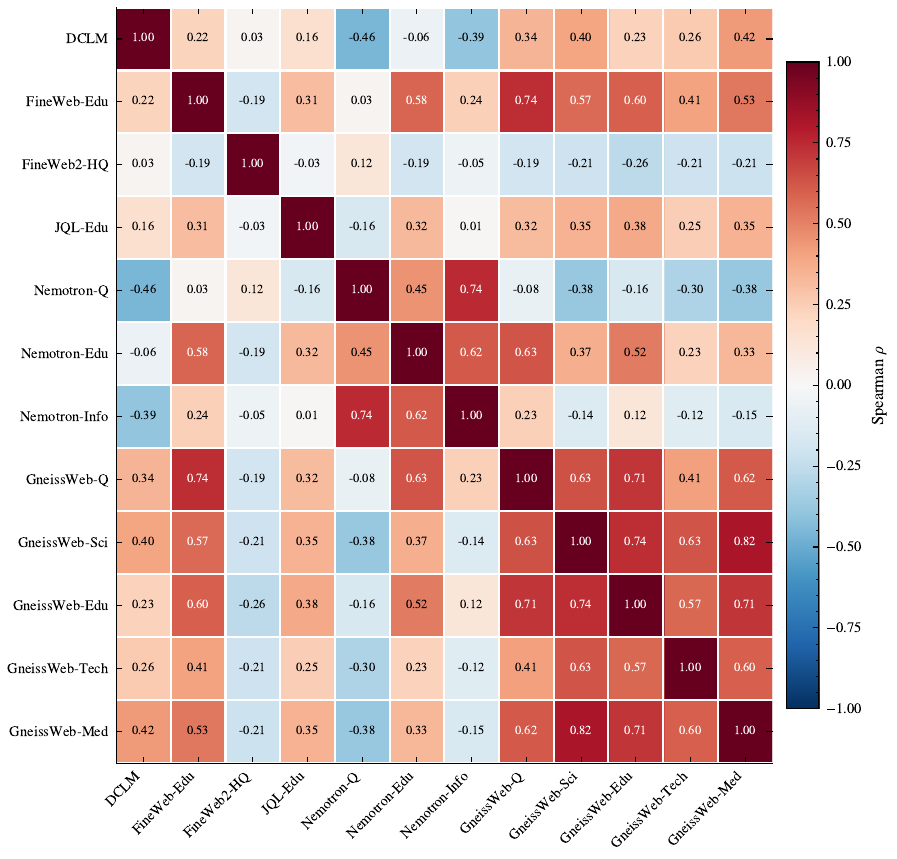}
\caption{Pairwise Spearman rank correlation between the twelve classifiers' document scores
(\valCoreDocs{} core documents; the diagonal is $1$ by construction). The scale runs from $-1$  %
(blue) through $0$ (white) to $+1$ (red). \DCLM is largely orthogonal to the other quality
signals and \emph{negative} against the Nemotron heads, whereas the GneissWeb topic scorers and
the Nemotron heads each form positively correlated blocks. This is the full pairwise view behind
the single-classifier robustness of the coverage result (Figure~\ref{fig:rq5rob}).}
\label{fig:clf_corr}
\end{figure*}

\paragraph{How the classifiers relate.} Figure~\ref{fig:clf_corr} gives the pairwise Spearman
correlation between the twelve classifiers' scores across the \valCoreDocs{} core documents  %
(every document scored by all classifiers). It contextualises two threads in the body. \DCLM
is largely orthogonal to the other quality signals---weakly correlated with most and
\emph{negative} against the Nemotron heads ($\valDCLMnemQ$ with Nemotron-Q)---consistent with  %
its outlier behaviour throughout (\S\ref{sec:rq2}) and its lone non-agreement in the
coverage-robustness check (Figure~\ref{fig:rq5rob}). The GneissWeb topic scorers and the
Nemotron heads each form positively correlated blocks, and \FWEdu aligns most closely with
GneissWeb-Q, so the general-purpose ``quality'' classifiers largely share a common signal that
\DCLM does not; FineWeb2-HQ, the multilingual model, correlates weakly or negatively with the
English-trained ones.

\end{document}